\documentclass[conference]{IEEEtran}
\IEEEoverridecommandlockouts
\usepackage{amsmath,amssymb,amsfonts}
\usepackage{algorithmic}
\usepackage{graphicx}
\usepackage{textcomp}
\usepackage{xcolor}
\usepackage{ragged2e}
\usepackage{cite}

\begin{document}

\title{
Feasibility of Neural Radiance Fields for Crime Scene Video Reconstruction\\}

\author{
\IEEEauthorblockN{Shariq Nadeem Malik}
\IEEEauthorblockA{\textit{Independent Researcher} \\
shariqnadeem581@gmail.com}
\and
\IEEEauthorblockN{Chee Min Hao}
\IEEEauthorblockA{\textit{School of IT} \\
\textit{Monash University Malaysia}\\
Petaling Jaya, Malaysia \\
mche0121@student.monash.edu}
\and
\IEEEauthorblockN{Dayan Mario Anthony Perera}
\IEEEauthorblockA{\textit{School of IT} \\
\textit{Monash University Malaysia}\\
Petaling Jaya, Malaysia \\
dper0033@student.monash.edu}
\and
\IEEEauthorblockN{Dr Lim Chern Hong}
\IEEEauthorblockA{\textit{School of IT} \\
\textit{Monash University Malaysia}\\
Petaling Jaya, Malaysia \\
lim.chernhong@monash.edu}
}

\maketitle

\begin{abstract}
This paper aims to review and determine the feasibility of using variations of NeRF models in order to reconstruct crime scenes given input videos of the scene. We focus on three main innovations of NeRF when it comes to reconstructing crime scenes -- Multi-object Synthesis, Deformable Synthesis, and Lighting. From there, we analyse its innovation progress against the requirements to be met in order to be able to reconstruct crime scenes with given videos of such scenes.
\end{abstract}

\section{Introduction}
The development of NeRF in the recent years has opened up the possibility of using it in more areas of forensic science such as  crime scene investigation and reconstruction. This review paper examines the feasibility of using the improvements made over the base NeRF model for enhancing crime scene analysis and novel view synthesis. 

\subsection{What is NeRF}
NeRF introduced by Mildenhall et al. in 2020 a method that optimizes a deep fully-connected neural network to represent the function of radiance accumulation by a ray passing through a 5D coordinate(x, y, z, theta, phi) to determine volume density and view dependant RGB color. It allows the generation of novel views from a sparse set of input images. 
It relies on MLP (multilayer perceptron to model radiance field with the goal being to create a continuous 5D radiance field from a representation of a scene from a set of input images)
NeRF (Neural Radiance Fields) works by representing a scene as a continuous 5D function that takes a 3D spatial location and a 2D viewing direction (represented by theta and phi) as input and outputs the radiance (color and opacity) at that point.
This function is parameterized by a deep neural network, which is trained to predict the radiance at any point in the scene given the input coordinates.

\subsection{The Dataset}
The dataset that has to be kept in context for crime scene investigation is the UCF Crime Dataset which was developed by Sultani et al \cite{sultani2018real} . It consists of videos divided into 13 categories (such as Abuse, Arrest etc). The key feature to observe here, is the fact that the videos in most cases contain only one angle of CCTV footage. This is where novel view synthesis can play a huge role in improving crime scene analysis as it can generate novel views which can aid investigations greatly. 

\section{Current Key Innovations in NeRF}
In this section we will be taking a look at the key innovations that make it possible to implement NeRF models. 

\subsection{Multi Object Synthesis}
Neural Radiance Fields (NeRF) models have shown promising results in novel view synthesis, but face challenges when dealing with complex scenes and multiple objects. Several recent works have addressed these limitations:

NeRF-MS tackles the challenge of leveraging multiple sequences of images in different conditions to improve upon the performance of current NeRF models on generating novel views \cite{10376583}. The paper addresses two main problems:

\textbf{Appearance variance.} Due to different lightning conditions, the NeRF models tend to overfit on the training data which results in inconsistent results and loss of complex textures" \cite{10376583}.

\textbf{Inclusion of non static objects (objects within a scene that are in motion) in the novel view synthesis.} With the existing NeRF models at the time, the issue was that the reconstruction of such objects was inconsistent and lead to artifacts" \cite{10376583}.

To address these issues, NeRF-MS introduces several key innovations:

\textbf{Introduction of a new Loss function.} To address the overfitting problem of 'per-image appearance codes', the authors introduced a triplet loss. The triplet loss function maintains multi-view consistency by maintaining similarity between appearance codes in the same scene and diversity between sequences" \cite{10376583}.

\textbf{Transient Decomposition Module.} Before looking into this specifically lets first get an idea of what 'transient object' means. Transient object/s refers to all objects which are not present in all frames of a single scene. We can also call them moving objects example: cars or people passing by" \cite{10376583}.

By using TDM, it helps NeRF distinguish between what is a static object (object that will stay throughout the scene) and what is a non static object (wont always be in the scene). This is done in two ways:

\textbf{Image transient codes.} These help identify objects that appear in individual images but aren't part of the permanent scene

\textbf{Sequence transient codes.} These help track objects that might appear in multiple images but aren't permanent features of the scene. \cite{10376583}

Similarly, NeRF in the Wild (NeRF-W) addresses scene reconstruction from unstructured photo collections and overcomes issues that rise from usage of structure-from-motion and image-based rendering for scene reconstruction \cite{martin2021nerf}. The paper discusses how moving objects cause a noticeable fall in the novel view synthesis. Proposed model in this paper tackles the issue which traditional models face where the assumption for the input is that they contain static objects. It also causes artifacts to appear and this is especially noticed when there is even little variation in the images (such as daylight, exposure etc) \cite{martin2021nerf}.

NeRF-W is a model built upon the base NeRF model. The inputs of this model are trained using unstructured collections of data. This helps in better synthesis of the novel views of the complex scenes" \cite{martin2021nerf}.

In terms of depth estimation and reflective surfaces, recent models have made significant progress:

\textbf{ref-NeRF} focuses on resolving the problem of recreating and modelling scenes which contain reflective surfaces by parameterizing NeRF radiance based on the reflection of the viewing direction about the local normal vector" \cite{tewari2022advances}.

\textbf{DS-NeRF} Utilizes depth supervision from point clouds extracted using COLMAP to enhance training efficiency and convergence speed by modeling depth as a normal distribution around the sparse point cloud depth" \cite{tewari2022advances}.

\textbf{COLMAP} (Structure-from-Motion and Multi-View Stereo) is a software package commonly used in Computer Vision for reconstructing 3D models from 2D images. It works by analyzing multiple images of a scene to estimate camera poses and scene structure, creating a sparse point cloud and dense depth maps" \cite{tewari2022advances}.

In the context of NeRF and depth supervision, COLMAP plays a crucial role in extracting sparse point clouds from training images. These sparse point clouds provide depth information that can be used to supervise the training of NeRF models, improving the accuracy of depth estimation and enhancing the overall quality of the rendered scenes" \cite{tewari2022advances}.

Depth estimation has been one of the most highlighted problems of the baseline NeRF model. In order to solve this problem, a new model has been created thanks to the work of Kengle et al in 2021\cite{deng2021depthsupervised}. 

This model by \cite{deng2021depthsupervised}, like the ones we discussed earlier, also addresses the novel view synthesis and multi object synthesis problem. The approach however, also brings in an added advantage of reducing the number of inputs required for the NeRF model to perform novel view synthesis with quality similar to the existing NeRF models which use the sparse-view approach.

In addition to this, it also accelerates the model training time by 2-3 times.

The DS-NeRF model’s depth-supervision loss also enables training of the model for handling scenes which have multiple objects or complex geometries.  

Using depth information to directly supervise the NeRF density function, DS-NeRF is able to more accurately capture the geometry and appearance of various objects in scene. It serves as a valuable signal in helping the model in understanding the different features of objects within a scene much more accurately. 
The technique can be used in conjunction with other depth sources, including depth cameras, to improve the synthesis of scenes with several objects. Using the depth information from several similar sources, it can help in introducing better accuracy and detail in the rendered scenes.

These advancements demonstrate the ongoing evolution of NeRF models to handle increasingly complex scenes and multi-object synthesis.

\subsection{Deformable Synthesis}
Deformable synthesis is a relatively unexplored area when it comes to Neural radiance fields particularly in the case of dynamic scenes consisting of multiple objects. The approach in this case, relates to the free-viewpoint synthesis which is a particularly challenging area when it comes to scene reconstruction. The primary issue with dynamic data is highlighted when Song et al. mentions that dynamic videos usually contain large-scene motions and the deformations need to be such that free-viewpoint synthesis is possible \cite{song_yang_deng_zhu_ramanan_2023}. This combined with having multiple actors and a multitude of interactions on the video make the task all the more challenging. 

Song et al. takes a unique approach when they deform the scene into object-centric representations which are further decomposed into global and local movements. This approach allows the reconstruction of videos upto a minute long and the object-centric approach makes it easy to extract the root-body motion of an object which is a key component in free-viewpoint synthesis. Furthermore, only a single camera is required compared to most other methods which have an elaborate setup and unknowingly forgo the real-world application of the overall task. However, the data used is RGBD which might not be a format that is common with most cameras and datasets, the use of an iPhone camera is also a problem as the quality is relatively too high for crime-scene reconstruction which generally sees lower-quality and far rougher videos. 

The problem of dynamic scenes is also taken up by Lu et al. when they try and make use of 3D sparse convolution to account for consistency in local information which is similar to Song et al. as they also account for local information in their network\cite{lu20243dgeometryawaredeformablegaussian}. They also use continous 6D rotation in order to accurately represent the rotational states of each gaussian at different time. Overall the architecuture consists of a gaussian canonical field which learns to reconstruct static scenes and a deformation field which learns object deformation and finally renders the result as a dynamic scene. 

Another interesting but similar problem is taken up by Yang et al. when they mention the difficulty of capturing intricate details in scenes which are not handled by prior implicit methods. To that end, they make use of 3D Gaussian Splatting which combined with a deformation field, allows the model to reconstruct monocular scenes \cite{yang_gao_zhou_jiao_zhang_jin_2023}. Although this leads to higher rendering quality and rendering speed, the work makes use of a dataset with a wide variety of perspectives which may not occur in the wild thereby causing the effectiveness of the model to be called into question like many other works regarding Neural Radiance Fields. Furthermore the efficiency of the method depending on pose estimations is another issue that will cause the usefulness of the model in crime-scene detection to be called into question as the datasets used do not reflect real-life scenarios. 

Das et al. tries to solve this common issue in their work where they use a two-stage approach that involves fitting a low-rank neural deformation model that learns object deformations and afterwards uses a coarse model to complete the reconstruction \cite{das2024neuralparametricgaussiansmonocular}. However the work still suffers due to further dataset limitations as scenes with complex movements eg. unsteady camera work remain untested. 

\subsection{Lighting}

NeRF in itself is difficult to render proper lighting for objects since the model learns with prior illumination and reflectance values with a known light source. Hence, various works within this area tend to focus on being able to render lighting for objects given input images from ambiguous light sources. Some works also enhanced shadowing and allow relighting of the object in order to make things more realistic, especially across different scenes.

Being one of the earlier works in this specific field, Boss et al.'s work focused on tackling the problem where NeRF is only able to generate RGB values during synthesizing, which makes it unable to cater to ambiguous lighting conditions. Therefore, the NeRD calculates the outgoing radiance of the object during decomposition which consists of the geometry, BDRF (Bidirectional Reflectance Distribution Function) values instead of RGB values, and incident illumination which allows for quick physical rendering and re-rendering when given different types of scenes. The final rendered object is able to illuminate according to any scene with different types of lighting conditions \cite{boss2021nerd}. This model is able to render these objects in real time given a set of similar pictures taken in different scenes with different illumination, which allows cross-checking of evidence under various lighting conditions. However, the model does not take account into shadowing and potential indirect illumination which is crucial in better mimic the object's placement and pointed direction in the crime scene.

Srinivasan et al. were also able to solve this problem but in a similar but also slightly different manner. In their work, although the sampling model generates a similar set of outputs and material properties from a set of similar images, the model specifically takes account of indirect illumination and not just direct illumination. Furthermore, since the output consists of an approximated shadow map, the physically-rendered object can be viewed from all points in the hemisphere with appropriate lighting and shadowing. This complements the lack of shadowing rendered from the previous work and also allows any point of view of the object in which it will render its illumination and shadow depending on the viewpoint \cite{nerv2021}. This makes it crucial in better estimating the time of day that the crime scene had happened, while also identifying certain key details of the crime scene where lighting on objects is deemed important, especially when it comes to placement of objects. However, we can see that crime scenes often consists of multiple objects instead of a single one. This is where our next model comes into play.

Wizadwongsa et al.'s work focuses on an interesting direction where it renders novel view synthesis with MPIs (Multi Plane Images), meaning a real-time scene is rendered from inputs consisting of various camera angles taken on a scene with multiple objects. The approach parameterize each pixel as a linear combination of learnable basis functions which makes rendering these generated parameters more efficient compared to fixed basis functions. The parameters generated are then rendered into view-dependent effects, mainly consisting of lighting. Finally, the finer details are rendered using a hybrid implicit-explicit modelling strategy, where it utilises the base color of the pixel to ease the compressing and reproducing of finer details of the modelled coefficient images. \cite{Wizadwongsa2021NeX} The paper emphasizes on how his work is able to clearly render colour-based lighting reflection (e.g. from a CD) which is something that the previous two models did not make tests on, which enables more types of objects to be rendered, including for a crime scene. However, MLPs usually only covers the hemisphere viewpoints in front of the scene and does not render the scene as a whole similar to objects rendered from both Srinisavan et al. and Boss et al.'s models, meaning that if it were to be used for a crime scene novel view synthesis, it would not be able to render the whole scene from front to back.

\vspace{12pt}

\begin{table}[ht]
\centering
\footnotesize
\begin{tabular}{|p{1cm}|c|c|c|p{1.5cm}|}
\hline
\textbf{Model} & \textbf{PSNR} & \textbf{LPIPS} & \textbf{SSIM} & \textbf{Dataset} \\
\hline
DS-NeRF & 18.1 & 0.40 & 0.62 & Redwood-3dscan (2-view) \\
\hline
DS-NeRF w/ RGB-D & 20.3 & 0.36 & 0.73 & Redwood-3dscan (2-view) \\
\hline
NeRF-W & 22.23--29.08 & 0.849--0.962 & 0.110--0.250 & Phototourism \\
\hline
NeRF-MS & 22.839 & 0.2347 & 0.7933 & NeRF-OSR \\
\hline
\cite{yang_gao_zhou_jiao_zhang_jin_2023} & 24.11 & 0.1769 & 0.8525 & D-NeRF+ HyperNeRF \\
\hline
\cite{das2024neuralparametricgaussiansmonocular} & 22.348 & 0.095 & 0.905 & D-NeRF+ HyperNeRF \\
\hline
Total-Recon & -- & 0.278 & -- & Custom \cite{song_yang_deng_zhu_ramanan_2023} \\
\hline
NeX & 26.45 & 0.165 & 0.890 & Shiny Dataset \\
\hline
NeRD & 23.86 & - & 0.88 & dataset\cite{boss2021nerd}\\
\hline
NeRV & 23.16 & - & 0.883 & lego\cite{nerv2021}\\
\hline
\end{tabular}
\hfill
\caption{\centering Performance Metrics of Different Models on Various Datasets}
\end{table}

\section{Current Progress on the implementation of NeRF models for Crime Scene Analysis}
There are currently many approaches taken in Neural Radiance Fields with respect to dynamic scene reconstruction however, direct work on crime-scene analysis has seen no light. Analysis of the work on dynamic scene reconstruction however can reveal insights as to the feasibility of the method. 

Through this although work has progressed a fair amount, the issue of NeRF's working with datasets that are comprised of data purely from the wild and not perfect has yet to take a step forward, which is a significant part of crime-scene analysis, particularly reconstruction. Taking purely the models, approaches that focus on the objects themselves have shown good results combined with methods that focus on extracting object geometry.

For lighting, it seems that there is work yet to be done as there is no work that is currently able to merge both MLP novel view synthesis with a full object view synthesis that takes account to various types of illumination, allowing the whole scene to be viewed from any point of view with proper relighting done. Furthermore, in Srinisavan et al., we can see that some part of the objects can be relighted through indirect illumination, which can significantly aid in rendering more realistic objects in a crime scene

Overall, work on crime-scene analysis has seen no progress in the area of NeRF's but through analysis of related areas like multi-object and dynamic scenes coupled with wild datasets we can make some inference regarding the feasibility of the study. 

\section{Conclusion}
Through this study, we were able to identify key innovations for novel view synthesis with the aim to reconstruct crime scenes to aid in crime investigations. We then also took these innovations and analysed its collective feasibility of being able to fully reconstruct a crime scene given its set of input videos, which we concluded that such a task has not been feasible yet to be achieved. However, we can see promising progress from our reviewed papers which makes this goal achievable in the long term.

\bibliographystyle{IEEEtran}
\bibliography{reference}

\end{document}